\documentclass[10pt,twocolumn,letterpaper]{article}

\usepackage[pagenumbers]{cvpr} 

\usepackage{graphicx, amsfonts, amsmath, amssymb, booktabs, caption, subcaption, multirow, overpic, textpos, cuted, bm, float}

\usepackage[table]{xcolor}
\usepackage[british, english, american]{babel}
\usepackage[pagebackref=false, breaklinks=true, letterpaper=true, colorlinks,
            citecolor=citecolor, linkcolor=linkcolor, bookmarks=false]{hyperref}
\definecolor{citecolor}{HTML}{0071BC}
\definecolor{linkcolor}{HTML}{ED1C24}

\usepackage[capitalize]{cleveref}
\crefname{section}{Sec.}{Secs.}
\Crefname{section}{Section}{Sections}
\Crefname{table}{Table}{Tables}
\crefname{table}{Tab.}{Tabs.}
\newcommand \myvec[1]{\mathbf{#1}}

\def\cvprPaperID{**}
\def\confName{****}
\def\confYear{****}

\newlength\savewidth

\renewcommand{\paragraph}[1]{\vspace{1.25mm}\noindent\textbf{#1}}

\newcolumntype{x}[1]{>{\centering\arraybackslash}p{#1pt}}
\newcolumntype{y}[1]{>{\raggedright\arraybackslash}p{#1pt}}
\newcolumntype{z}[1]{>{\raggedleft\arraybackslash}p{#1pt}}

\newcommand{\app}{\raise.17ex\hbox{$\scriptstyle\sim$}}

\definecolor{deemph}{gray}{0.6}

\definecolor{baselinecolor}{gray}{.9}

\newcommand{\authorskip}{\hspace{2.5mm}}

\begin{document}
\title{
\Large DFA-NeRF: Personalized Talking Head Generation via Disentangled Face Attributes Neural Rendering}
\author{
 Shunyu Yao \authorskip RuiZhe Zhong \authorskip Yichao Yan \authorskip
 Guangtao Zhai \authorskip Xiaokang Yang \\[2mm]
 Shanghai Jiao Tong University  \vspace{4mm}
}
\maketitle


\begin{abstract}
While recent advances in deep neural networks have made it possible to render high-quality images, generating photo-realistic and personalized talking head remains challenging. With given audio, the key to tackling this task is synchronizing lip movement and simultaneously generating personalized attributes like head movement and eye blink. 
In this work, we observe that the input audio is highly correlated to lip motion while less correlated to other personalized attributes (e.g., head movements).
Inspired by this, we propose a novel framework based on neural radiance field to pursue high-fidelity and personalized talking head generation. Specifically, neural radiance field takes lip movements features and personalized attributes as two disentangled conditions, where lip movements are directly predicted from the audio inputs to achieve lip-synchronized generation.
In the meanwhile, personalized attributes are sampled from a probabilistic model, where we design a Transformer-based variational autoencoder sampled from Gaussian Process to learn plausible and natural-looking head pose and eye blink. Experiments on several benchmarks demonstrate that our method achieves significantly better results than state-of-the-art methods.
\end{abstract}

\section{Introduction}
\label{sec:intro}


Synthesizing dynamic talking head driven by audio is of great importance to various applications, such as film production and online meeting. However, generating photo-realistic and expressive talking heads remains an open challenge, which not only contain accurate lip motions, but also present personalized eye blinks and head movements.

Traditional talking head generation models~\cite{Photorealistic, Hierarchical-Cross-Modal-Talking,wang2010synthesizing,song2018talking,zhu2018high,DBLP:conf/cvpr/CudeiroBLRB19} focus on synthesizing audio-synchronized lip motion, but only generate lip motion with fixed head poses. To address this issue, some recent works consider personalized attributes~\cite{yi2020audio, HDTF, MakeItTalk, wu2021imitating,wang2021anyonenet, Talking-head-Generation-with-Rhythmic-Head-Motion}. However, these methods~\cite{yi2020audio, HDTF, MakeItTalk} generate personalized information with a deterministic model and the results are short of diversity, leading to a repetitive pattern. Most of works utilize Generative Adversarial Networks (GAN)~\cite{Talking-head-Generation-with-Rhythmic-Head-Motion, Hierarchical-Cross-Modal-Talking} to generate the final images. However, GAN is prone to fall into mode collapse and it can only generate images with a fixed resolution. Recently, Neural Radiance Field (NeRF)~\cite{Nerf} 
achieves to render high-quality talking head images with unlimited resolution~\cite{adnerf, 4D_avatar, lombardi2019neural}. Nevertheless, these works neglect the personalized face attributes and fail to  synchronize audio with lip motion accurately.

In this work, we propose a novel framework, which we name Disentangled Face Attributes Neural Radiance Field (DFA-NeRF), to capture synchronized lip motion and personalized attributes simultaneously. As shown in Fig.~\ref{fig:intro}, instead of directly feeding the audio into neural radiance field~\cite{adnerf}, we propose first to predict the lip motion and personalized attributes from audio and then optimize dynamic NeRF conditioning on these features to synthesize talking head images. In this way, lip motion and personalized attributes are regarded as two disentangled representations, which guarantees the generated images to be lip-synchronized and have natural movement. 

The remaining question is how to predict these two types of features from audios. Our critical insight is that lip motion is highly related to auditory phonetics, while the personalized information, such as head poses and eye blinks, is weakly associated with audio and varies from person to person. Therefore, we argue that the key to tackling this challenge lies in two folds. \textbf{1)} As audio is strongly correlated with lip motion, it is more appropriate to approximate the lip motion with a deterministic model. Inspired by the recent success of contrastive learning in audio-visual synchronization tasks~\cite{zhou2019talking, nagrani2018seeing}, we introduce a contrastive learning strategy to synchronize the audio feature with the feature of lip motion.
\textbf{2)} The personalized attributes such as head poses and eye blinks are random and probabilistic. In order to model the distribution of the attributes and generate long time series, we propose a probabilistic model named Transformer Variational Autoencoder (VAE). VAE~\cite{casale2018gaussian} can generate smooth output and map the data into Gaussian distribution, while the attention mechanism in Transformer~\cite{vaswani2017attention} helps VAE learn the long-time dependence for time series. Furthermore, we model the temporal dynamics in the Transformer-VAE with a Gaussian Process (GP). 
With the deterministic and probabilistic model, our framework can generate disentangled motion features that align well with the talking head, providing better conditions for neural rendering.

In summary, our contributions include \textbf{1)} We propose two disentangled conditions (\ie, lip motion and personalized attributes) for neural radiance field to generate high-fidelity and natural talking head. 
\textbf{2)} We propose a self-supervised learning method based on AutoEncoder to disentangle lip motion and personalized attributes. \textbf{3)} We design a deterministic model to synchronize audio and lip motion and a probabilistic model to generate personalized attributes. 
\textbf{4)} We conduct extensive experiments on several benchmarks, and results demonstrate our method outperforms the state-of-the-art methods in generating high-quality talking head.

\section{Related Work}\label{sec:related}

\noindent \textbf{Audio-Driven Talking Head Generation.}
Generating photo-realistic video portraits in line with any input audio stream has long been a popular research topic in computer graphics and vision \cite{fan2015photo,fan2016deep, chen2020comprises}. 
Some methods aim at finding out the exact correspondence between audio and frames~\cite{schreer2008real, Lip, zhou2019talking, kr2019towards, zhou2018visemenet, edwards2016jali, prajwal2020lip,sadoughi2019speech,DBLP:conf/sii/ShimbaSYL15}.  

However, these methods usually ignore head motion since it is hard to separate head posture from facial movement.
Some methods are based on the 3D face reconstruction algorithm~\cite{guo2018cnn,anderson2013expressive,wang2012high, deng2019accurate,DBLP:conf/cvpr/CudeiroBLRB19} and GAN~\cite{DBLP:conf/siggraph/ReisCM20,DBLP:conf/eccv/PumarolaAMSM18,yi2020audio,Lahiri_2021_CVPR}.
They estimate intermediate representations such as 3D face shapes~\cite{karras2017audio, thies2020neural, yi2020audio} or 2D landmarks \cite{zakharov2019few, wang2020mead} to assist the generation process.
Unfortunately, low-dimensional intermediate representation is not capable of describing dynamic face deformations. Unlike these methods, instead of directly using intermediate face representations, our method only extracts the 3DMM face expression parameters to disentangle the mouth features and eye blink features in an implicit way. Moreover, the disentangled mouth features will synchronize with audio features in a high-dimensional latent space.
Recently, the neural radiance field (NeRF)~\cite{Nerf,DBLP:conf/cvpr/PumarolaCPM21,srn,kellnhofer2021neural} has been widely applied in 3D-related tasks because it can accurately reproduce complex scenes with implicit neural representation. 
Some recent works \etal~\cite{adnerf,4D_avatar} leverage NeRF to represent faces with audio features as conditions. 
Zhou \etal~\cite{zhou2021pose} modularizes audio-visual representations by devising an implicit low-dimension pose code. 

Our framework is also built on NeRF. Instead of directly sending the audio to NeRF, we propose two disentangled representations to provide improved conditions.

\begin{figure*}[t]
    \centering
    \includegraphics[width=1.0\textwidth]{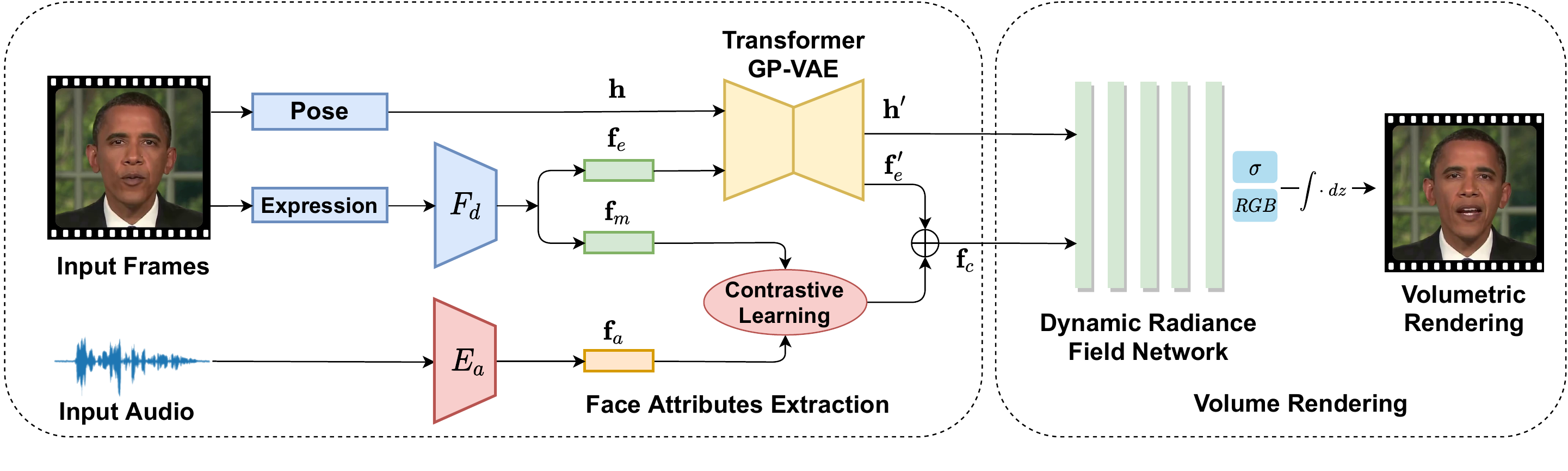}
    \caption{An overview of our proposed framework. Our method mainly consists of two parts: face attributes extraction and volume rendering. For the first part, pose and expression parameters are extracted from input videos. Then face attributes disentanglement module $F_d$ is introduced to disentangle eye blink embeddings $\myvec{f}_e$ and lip motion embeddings $\myvec{f}_m$. We use Transformer GP-VAE to generate personalized attributes such as head poses and eye blink features. We use a contrastive learning method for the lip motion to synchronize the audio feature $\myvec{f}_a$ with the mouth movement feature $\myvec{f}_m$. Afterward, in the volume rendering stage, we use generated pose $\myvec{h}^{\prime}$ as the view direction. Meanwhile, the generated eye blink feature and synchronized audio feature are concatenated and serve as a condition $\myvec{f}_c$ of the NeRF. Finally, we use volume rendering to render the image.}
    \label{fig:framework}
\end{figure*}

\begin{table*}[!t]
\footnotesize
    \centering
    \setlength{\tabcolsep}{2pt}
    \begin{tabular}{lccccccccc}
        \toprule
        Aspects & ATVG\cite{Hierarchical-Cross-Modal-Talking}& Wav2lip\cite{prajwal2020lip} & MakeItTalk\cite{MakeItTalk} & LSP\cite{Live-Speech-Portraits} & Yi \etal~\cite{yi2020audio} & Chen \etal~\cite{Talking-head-Generation-with-Rhythmic-Head-Motion} & FACIAL\cite{FACIAL} & AD-NeRF\cite{adnerf} & Ours \\
        \midrule
        Target  & arbitrary & arbitrary & arbitrary & specific  & specific & arbitrary & arbitrary & specific & specific       \\
        Audio feature & not sync   & sync & not sync & not sync & not sync & not sync  & not sync & not sync & sync\\
        Framework  & GAN & GAN & GAN & GAN & GAN & GAN & GAN  & NeRF & NeRF\\
        Personalized & No & No & heads, eyes & heads,eyes & heads, eyes & heads & heads, eyes & No & heads, eyes \\
        Face models& 2D landmarks & No & 2D landmarks& 2D landmarks   & 3DMM & 3DMM & 3DMM & No& No\\                                                                                    
        \bottomrule
    \end{tabular}
    \caption{We conclude five different aspects of several talking head generation works compared to our method: the methods work for a specific target or arbitrary images; The audio feature is synchronized with lip motions or not; The network architecture for image synthesis; The ability to generate personalized attributes and if they use any intermediate face models.}
    \label{tab:related_compare}
\end{table*}

\noindent \textbf{Personalized Face Attributes Generation.}
Personalized attributes are essential for generating natural talking head and have been studied in several prior methods. Yi \etal~\cite{yi2020audio} use LSTM ~\cite{LSTM} to process the input audio to generate expression and head pose. However, the generated head poses are predicted with a deterministic model, lacking diversity. MakeItTalk~\cite{AUTOVC} generates head pose through LSTM and MLP. However, for the LSTM network, the generation process is an auto-regressive task, making the head pose tend to \emph{freeze}. In FACIAL~\cite{FACIAL}, GAN is applied to learn attributes of the given audio and video, whereas the generation process is not explainable nor controllable. LSP~\cite{Live-Speech-Portraits} adopts a multi-dimensional Gaussian distribution to model the probabilistic process of head poses and body motions. The problem is that the distribution must be learned from scratch for different targets.

We propose a Transformer GP-VAE to generate personalized attributes, such as head poses and eye blinks. The Transformer architecture enables the model to generate long time series, and the VAE makes the generation process controllable and explainable. Furthermore, to model the temporal dynamics of the face attributes sequences, we adopt Gaussian Process to sample in VAE's latent space. The main differences between other methods and our approach are summarized in Table~\ref{tab:related_compare}.

\vspace{1mm}\section{Approach}\vspace{0.5mm}
\label{sec:approach}

\subsection{Overview}
As shown in Fig.~\ref{fig:framework}, given a source speech audio, our framework aims to construct natural talking heads. First, we extract the pose and 3DMM face expression parameters from video frames. We then design an AutoEncoder $F_d$ to disentangle the expression parameters into eye blink embeddings $\myvec{f}_e$ and mouth embeddings $\myvec{f}_m$ in a self-supervised manner. In the meantime, the input audio is sent to a CNN model to extract the audio features $\myvec{f}_a$. After that, we design a contrastive learning method to get the synchronized audio feature $\myvec{f}_{c}$. For head pose $\myvec{h}$ and eye blink embeddings $\myvec{f}_e$, a Transformer VAE with Gaussian Process is designed to model the probabilistic characteristic of these attributes. Finally, a dynamic NeRF is introduced to generate the final image with these conditions.

\subsection{Face Attributes Disentanglement}
\label{sec:exp}

We use 3D Morphable Face Models (3DMM)~\cite{3DMM} to extract face expression parameters. 
The 3D face landmark coordinates $\myvec{S}$ can be represented as the combination of 3DMM expression and geometry parameters:
    \begin{equation}
        \myvec{S} = \myvec{\bar{S}} + \myvec{B}_{id}\myvec{F}_{id} + \myvec{B}_{exp}\myvec{F}_{exp},
    \label{3dmm_face}
    \end{equation}
where $\bar{\myvec{S}} \in \mathbb{R}^{3N}$ is the averaged facial mesh, $\myvec{B}_{id}$ and $\myvec{B}_{exp}$ are the PCA basis of geometry and expression. $\myvec{F}_{id}$ and $\myvec{F}_{exp}$ are the coefficients of geometry and expression basis. Then we select 68 points in the facial mesh $\myvec{S}$ to get the face landmarks $\myvec{l}$ as in FaceWarehouse~\cite{FaceWarehouse}. 

\begin{figure}[t]
  \centering
   \includegraphics[width=0.9\linewidth]{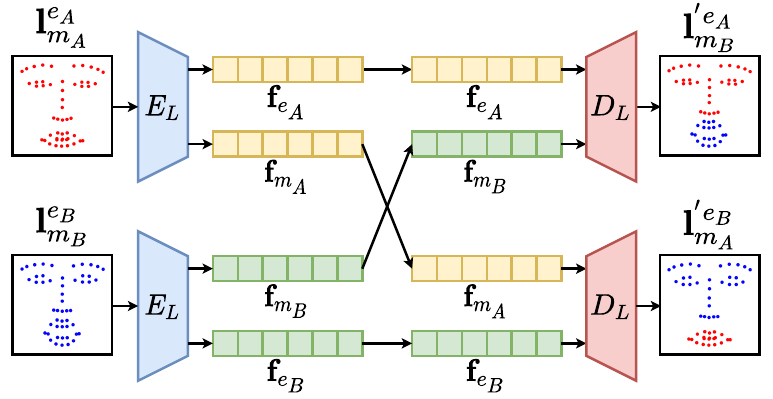}
   \caption{Illustration of face attributes disentanglement: the mouth embedding code is switched as an example. $\myvec{f}_e$ represents the eye blink embedding code, $\myvec{f}_m$ denotes the mouth embedding code. The subscripts $A$ and $B$ indicate different people.}
   \label{fig:face_dis}
\end{figure}

Whereas the expression code of 3DMM is entangled and unexplainable because it is based on PCA, which makes it difficult to control the movement of the mouth and eyes separately.
In order to generate disentangled face attributes, we propose a self-supervised learning method strategy. 
Specifically, we design a fully connected AutoEncoder, as shown in Fig.~\ref{fig:face_dis}. $\myvec{l}_{m_A}^{e_A}$ denotes the face landmarks of a target A. The subscript $m_A$ means the mouth landmarks of target A and $e_A$ denotes the eye landmarks. The AutoEncoder generates two intermediate embeddings $\myvec{f}_m$, $\myvec{f}_e$ with an encoder, which can be represented as $ \myvec{f}_{m_A}, \myvec{f}_{e_A} = E_L(\myvec{l}_{m_A}^{e_A})$. In the meantime, the decoder can be represented as $\myvec{l}_{m_A}^{' e_A}  = D_L(\myvec{f}_{m_A}, \myvec{f}_{e_A})$, which will be utilized to reconstruct the face landmarks with $L_1$ loss.
Given another face B and the corresponding face landmark $\myvec{l}_{m_B}^{e_B}$, we randomly switch the mouth embedding $\myvec{f}_m$ or eye blink embedding $\myvec{f}_e$ with their corresponding landmarks of face A and B. Afterwards, we can reconstruct different landmarks with:
\begin{equation}
    \mathcal{L}_{rec} = \lVert \myvec{l}_{m_B}^{' e_A} - \myvec{l}_{m_B}^{e_A} \rVert_1 + \lVert \myvec{l}_{m_A}^{' e_B} - \myvec{l}_{m_A}^{e_B} \rVert_1.\\
\label{eq:face_dis2}
\end{equation}
In this way, the AutoEncoder can reconstruct the face landmarks with different combinations of face attributes. Consequently, the embedding $\myvec{f}_e$ and $\myvec{f}_m$ can represent eyes and mouth movement in the learned latent space.

\subsection{Audio-Lip Synchronization}
After obtaining the mouth embedding $\myvec{f}_m$, we need to establish the relationship between the lip motion and the input audio. As indicated by previous works~\cite{prajwal2020lip, zhou2019talking}, learning the natural synchronization between visual mouth movements and auditory utterances is valuable for the talking face generation. Instead of using visual clues, we choose to synchronize the mouth movement embedding with the auditory utterances directly.
Specifically, we employ a CNN audio encoder $E_a$ to extract the phonemic feature $\myvec{f}_a$ from the input audio: $\myvec{f}_a = E_a(\myvec{a})$, where $\myvec{a}$ denotes the input audio data. We adopt a contrastive learning strategy to align audio features with mouth features to seek their synchronization.  
Specifically, we regard the timely aligned audio and mouth features $(\myvec{f}_a, \myvec{f}_m)$ as a positive pair, while the non-aligned pair $(\myvec{f}_a^{-}, \myvec{f}_m)$ is treated as negative.

We use the binary cross-entropy loss for contrastive learning, where the distance between timely aligned audio-mouth pairs should be closer than non-aligned pairs:
\begin{small}
    \begin{equation}
      \mathcal{L}_{con} = -\frac{1}{N} \sum^{N}_{i=1} y_i \log(d(\myvec{f}_m, \myvec{f}_a)) +
      (1 - y_i)\log(1 - d(\myvec{f}_m, \myvec{f}_a^{-})),
    \label{eq:contrastive}
    \end{equation}
\end{small}
where ${d}(\myvec{f}_a, \myvec{f}_m) = \frac{\myvec{f}_a \cdot \myvec{f}_m}{\lVert \myvec{f_a} \rVert_2 \cdot \lVert \myvec{f_m} \rVert_2}$ denotes the cosine distance, $y_i = 1$ for positive samples and $y_i = 0$ for negative samples and the total number of samples is $N$. During training, we randomly sample timely aligned audio and mouth sequences as positive pairs, while the negative pairs come from different videos or the same video with a time-shift.

\begin{figure}[t!]
    \centering
    \subfloat[Structure of the Transformer-VAE\label{subfig-1-1}]{
        \includegraphics[width=1.0\linewidth]{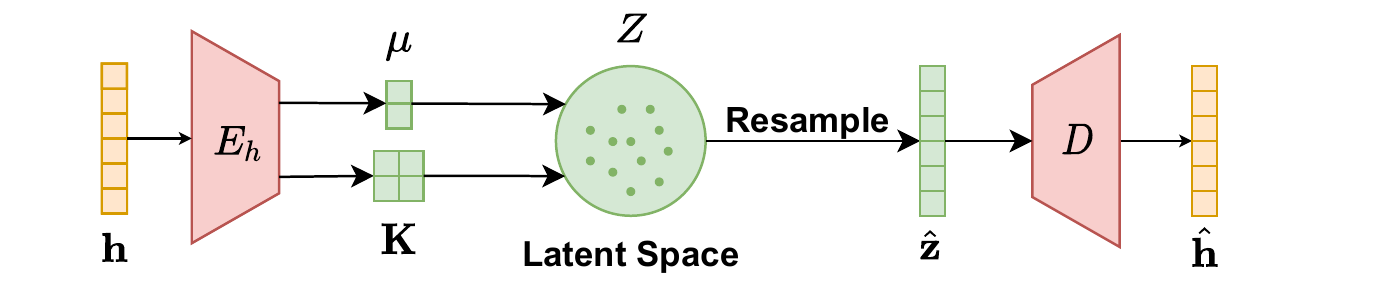}
    }\\
    \subfloat[Structure of the cross-modal encoder\label{subfig-1-2}]{
        \includegraphics[width=1.0\linewidth]{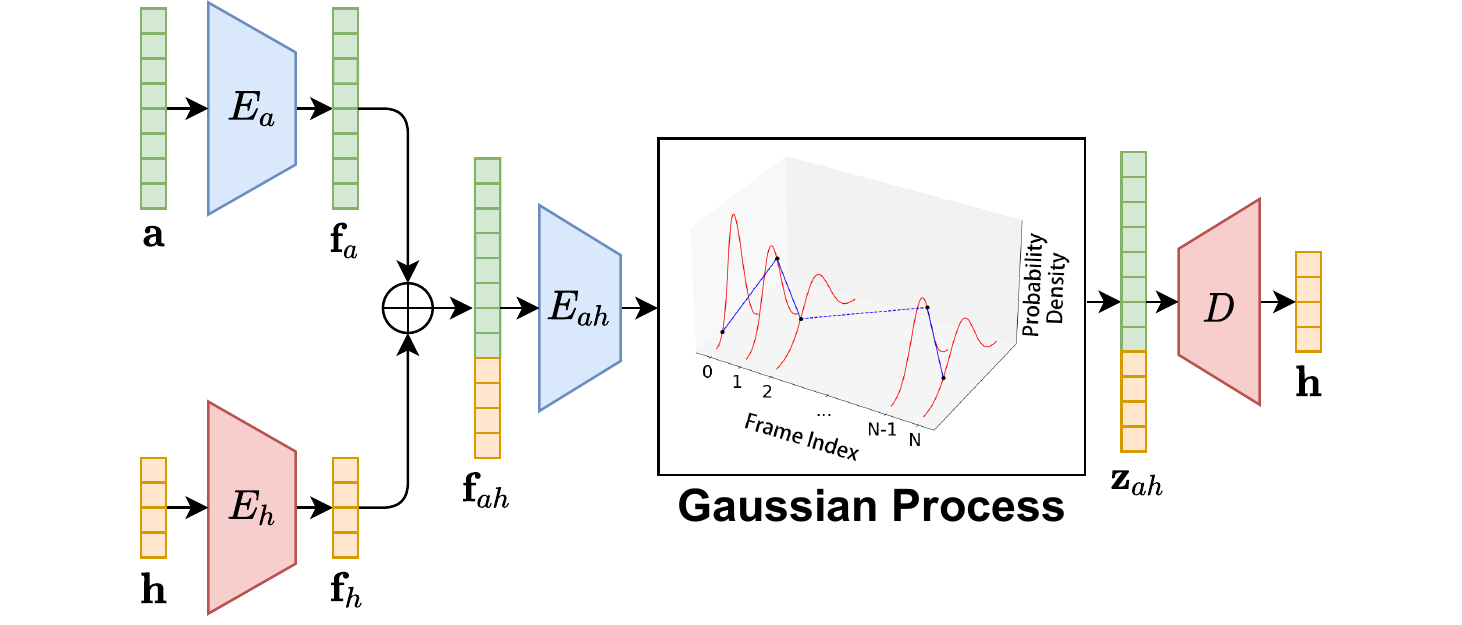}
    }
    \caption{a) Structure of Transformer VAE. This module will restructure the input face attribute sequences $\myvec{h}$ and learn a Gaussian prior latent space $Z$. The encoder $E_h$ and decoder $D$ are both Transformers. b) Structure of Transformer-based cross-modal encoder. Audio and BOP sequences share a common feature space, same as the trained latent space $Z$ in (a). We use the reparameterization trick to resample new latent codes $\myvec{z}_{ah}$ through Gaussian Process. The trained decoder $D$ decodes $\myvec{z}_{ah}$ to predict future face attributes.}
    \label{fig:VAE and Cross-modal encoder}
\end{figure}

\subsection{Personalized Attributes Generation}
To generate personalized attributes such as eye blinks and head poses, we propose a controllable probabilistic model named Transformer GP-VAE in Fig.~\ref{fig:VAE and Cross-modal encoder}, aiming to generate smooth and long time series. Given a face attribute (head poses or eye blinks) sequence $\myvec{h}_{1:T}$ with length $T$ and a more extended conditioning audio sequence $\myvec{a}_{1:T'}$ with length $T'$, we need to generate the face attribute embeddings $\myvec{h}_{T+1:T'}$ of the future $T' - T$ frames. Our method consists of two major parts. $\myvec{(1)}$ Latent space construction, where we train a Transformer-VAE with Gaussian Process (GP) on a large dataset to build a mapping between the input face attributes sequence and a latent space $Z$.
$\myvec{(2)}$ Face attributes generation, where we
fine-tune a cross-modal encoder on a selected person to embed both Beginning of Pose (BOP) and audio into the learned latent space $Z$. Thus, we can generate new face attributes sequences in an autoregressive manner.
Different from the original GP-VAE~\cite{GPVAE} and MGP-VAE~\cite{MGP-VAE}, our Transformer GP-VAE can generate various-length time series with cross-modal input.

\noindent \textbf{Latent Space Construction.} Assume our dataset contains $N$ continuous time series $\myvec{h}^{(i)}\in \mathbb{R}^{T \times d}$, where $T$ is the length of sequence and $d$ is the dimension of the vector. We construct VAE with a Transformer illustrated in Fig.~\ref{subfig-1-1}. We have also considered LSTM, but its outputs tend to converge to a fixed value and have no variations. The multi-head attention mechanism and fully-connected attention in Transformer tackle this frame-freezing problem. The detailed comparison experiments will be discussed in Section~\ref{sec:experments}.

\par Specifically, the VAE aims to map the time series $\myvec{h}$ to a multivariate normal distribution (latent space $Z$) with mean vector $\bm{\mu}$ and covariance matrix $\myvec{K}$:
\begin{equation}
        \bm{\mu} = \bm{\mu}_{\phi}(\myvec{h}),
        \myvec{K} = \myvec{K}_{\phi}(\myvec{h}),
    \label{eq:posterior-distribution-variable}
\end{equation} where $\phi$ are parameters of the encoder $E_h$ in VAE and the posterior distribution of $\myvec{z} \in Z$ is presented as:
\begin{equation}
    q_{\phi}(\myvec{z}\,|\,\myvec{h}) = \mathcal{N}\left( \bm{\mu}, \myvec{K}\right).
    \label{eq:posterior-distribution}
\end{equation}
Then we use the reparameterization trick to resample $\hat{\myvec{z}} \sim q_{\phi}(\myvec{z}\,|\,\myvec{h})$ and decode it into $\hat{\myvec{h}} = D(\hat{\myvec{z}})$. The Mean Square Error (MSE) with batch size $B$ is calculated to reconstruct $\hat{\myvec{h}}$:
\begin{equation}
    \mathcal{L}_{rec} = \frac{1}{B} \sum_{i=1}^{B} \lVert \hat{\myvec{h}}^{(i)} - \myvec{h}^{(i)} \rVert_2.
    \label{eq:MSE of reconstruction of h}
\end{equation}

\par Considering the prior distribution of $\myvec{z} \in Z$, the data points should be $N$ i.i.d. samples in the original VAE~\cite{VAE}. A time sequence does not satisfy such property due to the solid temporal correlation among frames. To address this, we utilize Gaussian Process to model the temporal correlation.
Given the time indices $\myvec{t}$ and the corresponding $\myvec{z}$ in $Z$ space, the prior distribution of $\myvec{z}$ is:

\begin{equation}
    p(\myvec{z}) = \mathcal{N}(\bm{\mu}(\myvec{t}), \myvec{K(t,t)}),
    \label{eq:prior distribution of z}
\end{equation}

where $\bm{\mu}(\myvec{t})$ and $\myvec{K}(\myvec{t},\myvec{t})$ are determined by the kernel function in GP, which we employ Cauchy kernel function~\cite{GPVAE}. KL divergence is calculated between prior and posterior distributions as another loss term:

\begin{equation}
    \mathcal{L}_{\mathrm{KL
    }} = \mathrm{KL}\left(p(\myvec{z}), q_{\phi}(\myvec{z}\,|\,\myvec{h}) \right).
    \label{eq:KL-divergence}
\end{equation}

After the VAE is trained, we can get the distribution of $\myvec{z}$ and the corresponding latent space $Z$. As we train the VAE with a large dataset, the model can generalize well because the latent space $Z$ contains different face attributes of the training sequences.

\begin{figure*}[t!]
    \centering
    \includegraphics[width=0.95\textwidth]{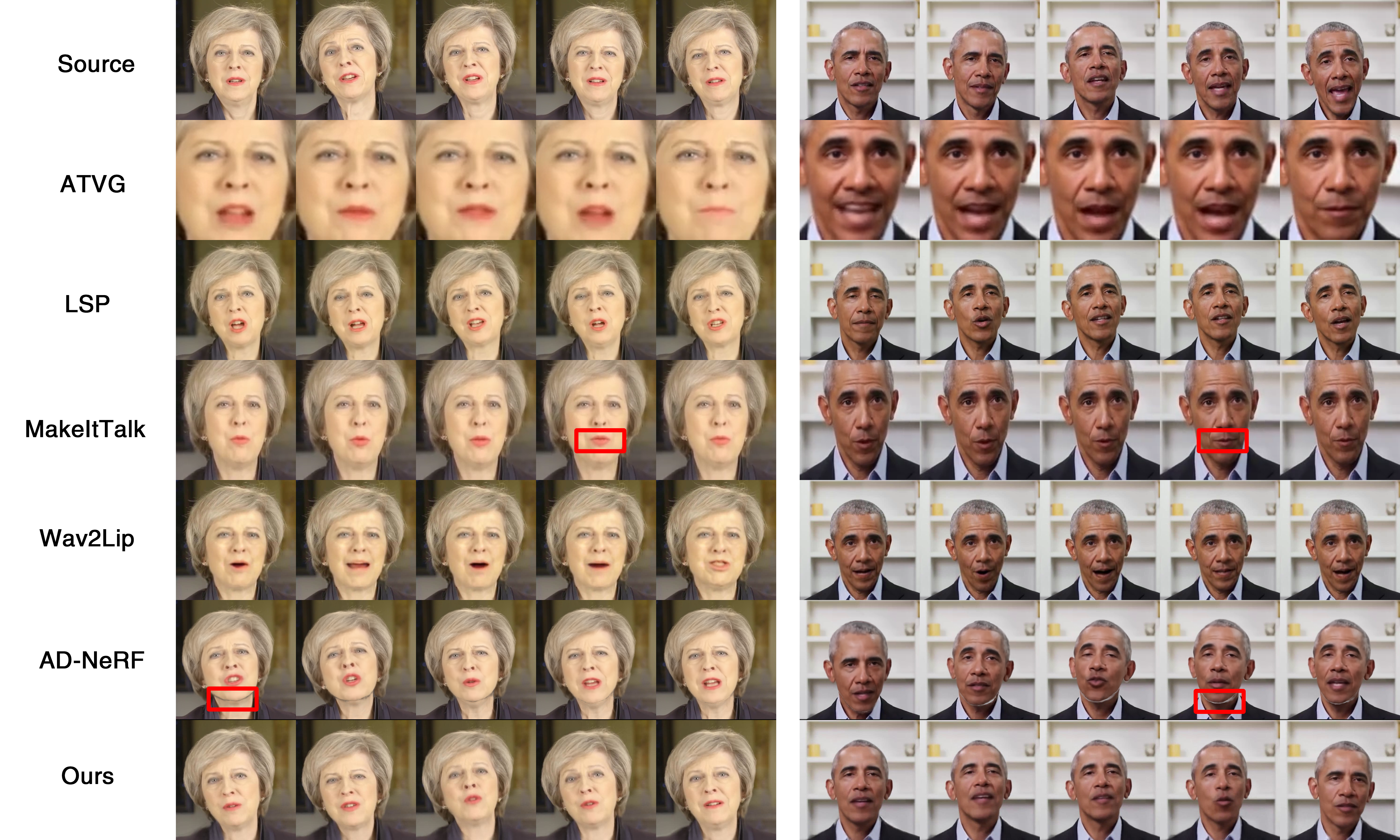}
    \caption{Qualitative results. The top rows are the reference source videos. ATVG~\cite{Hierarchical-Cross-Modal-Talking} can only generate the cropped frontal talking head. LSP~\cite{Live-Speech-Portraits} can generate natural head movements and eye blinks, but the lip motion is not accurately synchronized with the input voice. Though MakeitTalk~\cite{MakeItTalk} can achieve subtle head movements and eye blinks, the mouth shapes are not accurate, which are marked with a box. The lip motion of Wav2lip~\cite{prajwal2020lip} is accurate, but it can only synthesize the mouth movement of the talking face. AD-NeRF~\cite{adnerf} use two NeRFs to generate the head, lower body separately, and we observe a clear white gap between the head part and the body part(marked with a box in the figure). Our method generates photo-realistic talking heads with diverse head movements.}
    \label{fig:main_compare}
\end{figure*}

\noindent \textbf{Face Attributes Generation.} We fine-tune a cross-modal encoder on a selected person to generate personalized face attributes illustrated in Fig.~\ref{subfig-1-2}.
To generate novel face attributes, we need to encode both audio and BOP of a selected person into the trained latent space $Z$ in VAE, where audio serves as the conditional information and BOP is needed for initialization. Besides, the trained encoder $E_h$ and decoder $D$ in VAE will be used in this stage. 

\par For the audio information, we employ the same audio encoder $E_a$ which has been used in the audio-lip synchronization module to extract the input audio feature $\myvec{f}_{a}= E_a( \myvec{a}_{1:T} )$. In the meantime, we use the trained encoder ($E_h$) in Transformer VAE to encode the BOP to the trained latent space $Z$, \ie, $\myvec{f}_h  = E_h( \myvec{h}_{1:\tau} )$, where $\tau$ is the length of BOP sequence. Afterwards, these two features are concatenated in temporal dimension as $\myvec{f}_{ah} = \left[ \myvec{f}_a : \myvec{f}_{h}\right ]$ and fed into $E_{ah}$ to get the mean vector $\bm{\mu}^{\prime}$ and covariance matrix $\myvec{K}^{\prime}$ of cross-modal latent codes $\myvec{z}$'s distribution like $\myvec{z} \sim \mathcal{N}(\bm{\mu}^{\prime}, \myvec{K}^{\prime})$. We use the reparameterization trick to resample $\myvec{z}_{ah}$ from $\mathcal{N}(\bm{\mu}^{\prime}, \myvec{K}^{\prime})$ with Gaussian Process. Finally, the trained decoder ($D$) in Transformer VAE will decode $\myvec{z}_{ah}$ into the predicted time series $\hat{\myvec{h}}_{\tau+1:T}  = D ( \myvec{z}_{ah} )$.
The loss function of training cross-modal encoder can be calculated as:

\begin{equation}
    \mathcal{L}_{CME} = \frac{1}{B} \sum_{i=1}^{B} \left \Vert \, \hat{\myvec{h}}_{\tau+1:T}^{(i)} - \myvec{h}_{\tau+1:T}^{(i)} \, \right \Vert.
    \label{eq:cross-modal-encoder-loss}
\end{equation}

In the test phase, we feed the given audio and BOP to the cross-modal encoder to get the distribution of $\myvec{z}_{ah}$. By resampling $\myvec{z}_{ah}$ in this distribution and feeding $\myvec{z}_{ah}$ to decoder $D$, we can get the generated future face attributes. Note that we can also sample $\myvec{z}$ in the trained latent space $Z$ and feed it to the decoder to generate various face attributes without the input audio, which makes our Transformer-VAE more universal and controllable.

\subsection{Neural Scene Representation for Talking Head}
After obtaining the generated head poses, eye blink features $\myvec{f}_e$ and synchronized audio features $\myvec{f}_a$, we employ a neural radiance field to generate the final image with these conditions. We first concatenate audio embedding $\myvec{f}_a$ and eye blink embedding $\myvec{f}_e$ into a new embedding $\myvec{f}_c$. Then we present a conditional radiance field with this new embedding serving as input. After we transform the head pose from camera space into canonical space, we can directly use the head pose to replace the view direction $\myvec{d}$ of the radiance field. Finally, the embedding $\myvec{f}_c$, view direction $\myvec{d}$ and 3D location $\myvec{x}$ in canonical space constitute the input of the implicit function $F_\theta$. In practice, $F_\theta$ is realized by a multi-layer perceptron. With all concatenated input vectors, the network $F_\theta$ will estimate color values $\myvec{c}$ accompanied with densities $\sigma$ along with the dispatched rays. The entire implicit function can be formulated as:
\begin{equation}
   F_{\theta} : (\myvec{f}, \myvec{d}, \myvec{x}) \longrightarrow (\myvec{c}, \sigma).
\label{eq:nerf1}
\end{equation}
Then we can employ the volume rendering process by accumulating the sampled density $\sigma$ and RGB values $\myvec{c}$ along with the rays $r$ cast through each pixel to compute the output color $C(\myvec{r})$ for image rendering results. Finally, we use the photo-metric reconstruction error $\mathcal{L}_{photo}$ between the rendering outputs and the ground truth to train our NeRF: 
    \begin{equation}
        \mathcal{L}_{photo} = \sum_{\myvec{r}\in \mathcal{R}} {\lVert \hat{C}(\myvec{r}) - C(\myvec{r}) \rVert^2},
    \label{eq:nerf2}
    \end{equation}
Where $\mathcal{R}$ means the collections of the rays.  Please refer to supplementary material for more information on volume rendering and neural radiance field.

\begin{table*}[t!]
    \centering
    \setlength{\tabcolsep}{5pt}
    \begin{tabular}{lcccccccccc}
        \toprule
        ~ & \multicolumn{5}{c}{Testset A by \cite{Live-Speech-Portraits}} & \multicolumn{5}{c}{Testset B by \cite{Live-Speech-Portraits}}\\
        
        \cmidrule(lr){2-6} \cmidrule(lr){7-11}
        \text{Method} & $\text{PSNR} \uparrow$ & $\text{SSIM} \uparrow$ & $\text{LMD} \downarrow$& $\text{Sync} \uparrow$ & $\text{blinks/s}$ & $\text{PSNR} \uparrow$ & $\text{SSIM} \uparrow$ &$\text{LMD} \downarrow$ &$\text{Sync} \uparrow$ & $\text{blinks/s}$\\

        \midrule
        ATVG~\cite{Hierarchical-Cross-Modal-Talking} & 26.41 & 0.803 & 5.32 & 7.2 & N/A & 25.43 & 0.808 & 5.63 & 7.0 & N/A \\ 
        
        MakeitTalk~\cite{MakeItTalk} & 26.33 & 0.814 & 4.0 & 7.32 & 0.37 & 27.14 & 0.822 & 7.62 & 5.2  & 0.37 \\ 
        
        Wav2Lip~\cite{prajwal2020lip} & 27.63 & 0.854 & 5.63 & \textbf{9.8} & N/A & 27.32 & 0.851 & 5.45 & \textbf{9.5}  & N/A \\ 
        
        LSP~\cite{Live-Speech-Portraits} & 28.33 & 0.921 & 4.72 & 5.4 & 0.41 & 29.36 & 0.834 & 4.53 & 5.2  & 0.38 \\ 
        
        AD-NeRF~\cite{adnerf} & 28.72 & 0.914 & 5.42 & 4.3 & 0.12 & 31.12 & 0.903 & 4.86 & 4.5  & 0.84 \\ 
        
        \midrule
        Ground Truth & N/A & 1.000 & 0.00 & 9.4 & 0.42  & N/A & 1.000 & 0.00 & 7.6  & 0.32 \\ 
        DFA-NeRF (ours) & \textbf{29.12} & \textbf{0.938} & \textbf{3.74} & 7.3 & 0.46 & \textbf{32.09} & \textbf{0.917} & \textbf{3.84} & 7.1 & 0.37 \\ 
        
        \bottomrule
    \end{tabular}
    \caption{Evaluation with different methods. We use the videos \emph{May} and \emph{Obama2} as the testset A and testset B proposed in~\cite{Live-Speech-Portraits}. We train a new model with the method proposed in AD-NeRF~\cite{adnerf} and use the pre-trained models provided by all other methods to compare. We also report the eye blink frequency where N/A means the method fails to generate eye blink motions. Best results are in \textbf{bold}.}
    \label{tab:main_results}
\end{table*}

\begin{table*}[!t]
    \centering
    \setlength{\tabcolsep}{8.3pt}
    \begin{tabular}{lcccccc}
        \toprule
        MOS on / Approach & ATVG & MakeitTalk & Wav2lip & LSP & AD-NeRF & DFA-NeRF (ours)  \\ 
        \midrule
        Auido-Visual Sync & 3.42 & 2.18 & 3.76 & 3.66 & 3.65 & \textbf{3.84}  \\ 
        Naturalness of Head Movement & 1.32 & 2.43 & 1.47 & 4.03 & 3.63 & \textbf{4.12}  \\ 
        Naturalness of Eye Blink & 1.12 & 3.12 & 1.32 & 4.06 & 2.72 & \textbf{4.08}  \\ 
        Realness of Image& 1.52 & 2.24 & 2.53 & \textbf{3.84} & 3.63 & 3.78  \\ 
        \bottomrule
    \end{tabular}
    \caption{User study analyses measured by Mean Opinion Scores. Higher is better.}
    \label{tab:user_study}
\end{table*}
\section{Experiments}\label{sec:experments}
\subsection{Experimental Setup}

\noindent \textbf{Datasets.} We leverage the LRS2~\cite{chung2017lip} and HDTF~\cite{HDTF} datasets for the face attributes disentanglement and audio to lip synchronization tasks. LRS2 consists of thousands of spoken sentences from BBC television. Each sentence contains up to 100 characters. HDTF is a new dataset with high-resolution videos. It is collected from youtube and consists of about 16 hours of 720P or 1080P videos. In order to compare with previous SOTA methods~\cite{adnerf, Live-Speech-Portraits}, we also adopt the training videos in AD-Nerf~\cite{adnerf} and Live Speech Portraits~\cite{Live-Speech-Portraits}. 
For the LRS2 and HDTF datasets, we first estimate the 3DMM face expression parameters as in~\cite{deng2019accurate} and fix them to the front view. All the video data are resampled to 25 FPS. As for the audio data, we extract the MFCC audio feature with the windows size of 10 milliseconds and make sure that the audio sequences and mouth movements sequences are strictly aligned.

\noindent \textbf{Evaluation Metrics.} We conduct quantitative evaluations on metrics previously involved in the talking head generation field.
We employ Peak Signal-to-Noise Ratio (PSNR)~\cite{PSNR_SSIM} and Structural Similarity (SSIM)~\cite{PSNR_SSIM} to measure the image quality. Besides, we utilize both Landmarks Distance (LMD)~\cite{DBLP:conf/eccv/ChenLMDX18} around the mouths and the conﬁdence score (Sync$_{conf}$) proposed in SyncNet~\cite{casale2018gaussian} to evaluate the accuracy of mouth shapes and lip synchronization. Furthermore, we calculate the eye blink frequency (blinks/s) to evaluate the eye blink naturalness compared to the target person.

\subsection{Comparison with State-of-the-arts}
We compare our DFA-NeRF with the SOTA methods, including three identity-independent approaches, \ie, ATVG~\cite{Hierarchical-Cross-Modal-Talking}, MakeItTalk~\cite{MakeItTalk}, Wav2Lip~\cite{prajwal2020lip} and two identity-specific approaches, \ie AD-NeRF~\cite{adnerf} LSP~\cite{Live-Speech-Portraits}.

\noindent \textbf{Qualitative Results.} As subject evaluation is crucial for talking head generation, we show the generation results compared to the state-of-the-art methods. As Fig.~\ref{fig:main_compare} shows, our method can generate more realistic results and diverse head motions.
Only our approach and AD-NeRF can generate unlimited resolution videos among these methods.
AD-NeRF\cite{adnerf} uses two NeRF to render the head and torso separately and blend them. Therefore, it cannot avoid artifacts appearing at the neck. In contrast, we adopt one NeRF to simultaneously render the head and torso, consequently generating more appealing results.

\noindent \textbf{Quantitative Results.} The quantitative results are reported in Table~\ref{tab:main_results}. It can be observed that our method achieves the best results under most of the metrics on the test dataset. In Testset A, our model achieves 29.12 in PSNR and 0.938 in SSIM, notably outperforming prior arts. Our model also achieves the lowest LMD, which means it can predict the most accurate mouth shapes compared to other methods. As for the experiments on eye blink frequency, we observe that ATVG\cite{Hierarchical-Cross-Modal-Talking} and Wav2Lip\cite{prajwal2020lip} fail to generate the eye blink motions. The eye blink frequency of AD-NeRF\cite{adnerf} is low and the eyes cannot completely close. The eye blink frequency of MakeitTalk\cite{MakeItTalk} is fixed for the two testset. Only our method and LSP\cite{Live-Speech-Portraits} can generate personalized eye blinks naturally. The reason for the high Sync$_{conf}$ in Wav2Lip~\cite{wang2020mead} is that it only generates the lip movements of the input videos, and the Sync$_{conf}$ is even higher than the ground truth video. The Sync$_{conf}$ of our method is closest to the ground truth, which means the input audio and our generated lip motion are well synchronized.

\noindent \textbf{User Study.}  We conduct user studies with 20 attendees on 36 videos generated by ours and the five other methods. The driving audio is selected from three different languages: English, Chinese and German. We adopt the widely used Mean Opinion Scores (MOS)~\cite{DBLP:conf/ism/XuXPJ11} rating protocol. Each participant is asked to rate from 1-5 for the talking-head generation results based on four major aspects: audio-visual synchronization quality, the naturalness of head movement, the naturalness of eye blink and image realness. We collect the rating results and compute the average score of each method. The statistics are shown in Table~\ref{tab:user_study}. Most of the method has comparable Audio-Visual synchronized score except for MakeitTalk~\cite{MakeItTalk}. Wav2lip~\cite{prajwal2020lip} and ATVG~\cite{Hierarchical-Cross-Modal-Talking} cannot change head poses and generate eye blink motions. Therefore, their naturalness of head movement and eye blink scores are relatively lower. Although LSP~\cite{Live-Speech-Portraits} achieves the highest score of the image realness(3.84), our method achieves comparable results(3.78). Users prefer our results on audio-visual synchronization, the naturalness of head movement and eye blink, proving the effectiveness of the proposed method.  
\begin{figure}[t]
    \centering
    \includegraphics[width=1.0\linewidth]{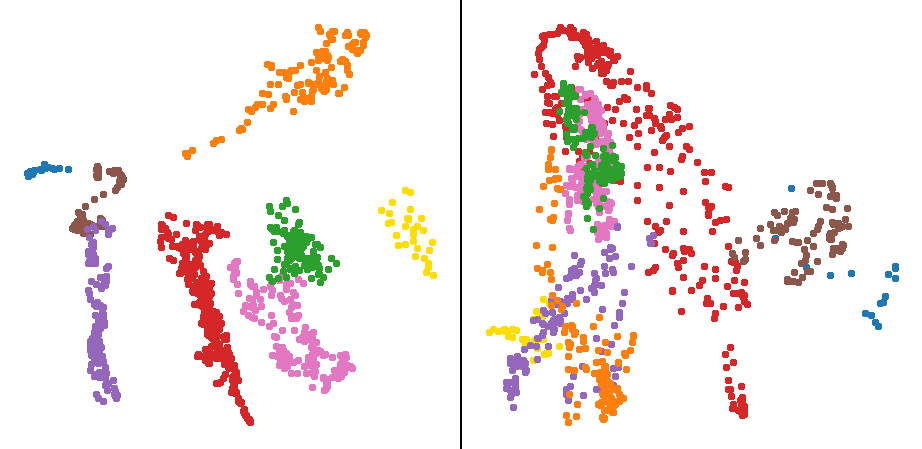}
    \caption{PCA of different people's personalized face attributes in $Z$ space. Left: with audio as the condition. Right: without audio as the condition. Different kinds of colors represent different identities. We find that the audio information helps the cross-modal encoder predict personalized face attributes.}
    \label{fig:z_pca_audio_bop}
\end{figure}
\subsection{Further Analysis} 
\noindent \textbf{Analysis of Personalized Attributes.} In our framework, we employ audio to predict the personalized attributes. To verify whether the audio information helps identify personal attributes, we visualize the PCA results of different people's personalized attributes in the $Z$ space with or without audio. Fig.~\ref{fig:z_pca_audio_bop} demonstrates that the personalized attributes are more distinct conditioned on audio. 

\begin{table}[t]
  \centering
  \begin{tabular}{lcc}
    \toprule
    Method & D-Rot$\,\,\downarrow$  & D-Pos$\,\,\downarrow$\\
    \midrule
    MLP VAE & 17.81\% & 20.95\% \\
    LSTM VAE & 19.06\% & 22.81\% \\
    Trans VAE & 3.45\% & 5.21\% \\
    Trans GP-VAE (Ours)& \textbf{3.19\%} & \textbf{4.47\%}\\
    \bottomrule
  \end{tabular}
  \caption{Ablation studies with the structure and sample process of VAE. Trans refers to Transformer.}
  \label{tab:head_pose_GPVAE_ablation_study}
\end{table}

\begin{figure}[t]
  \centering
   \includegraphics[width=0.9\linewidth]{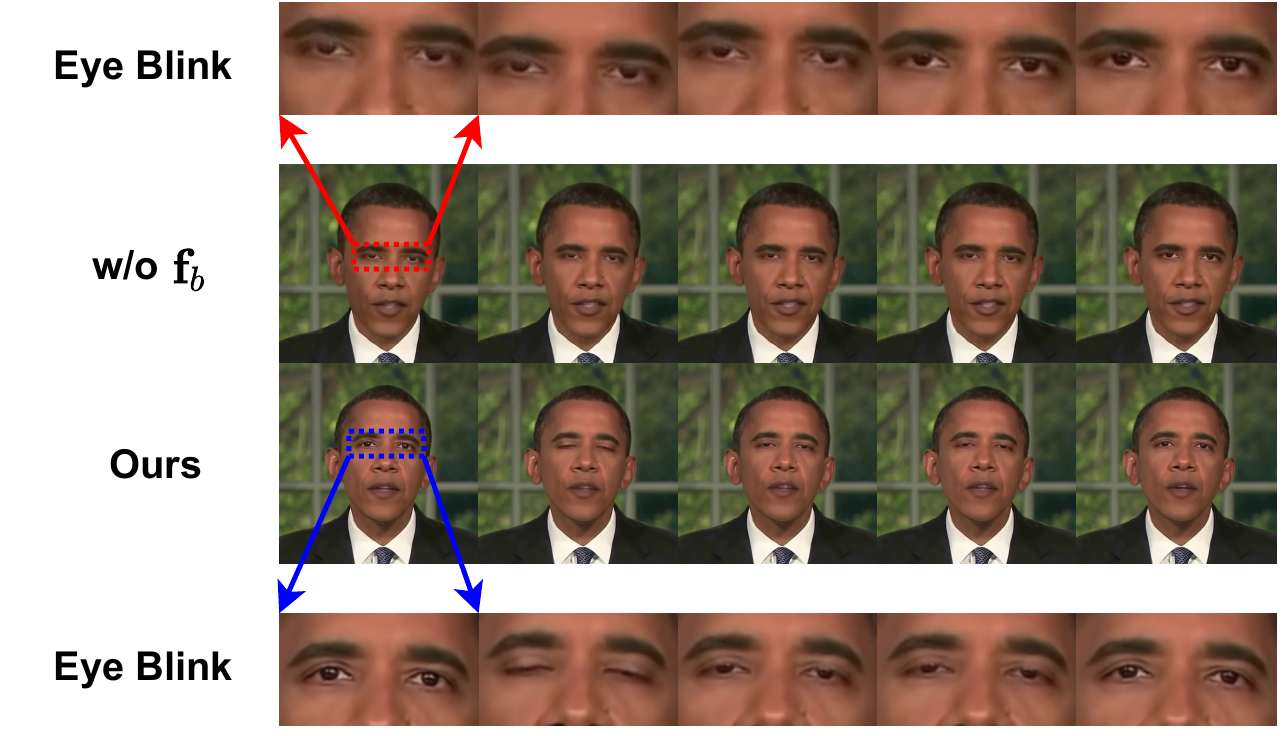}

   \caption{Ablation study of the disentangled eye blink features. We present continuous sequences of synthesized videos. The upper row reveals that the eyes fail to blink without the feature $\myvec{f}_e$. Our method can synthesize natural eye blink images with $\myvec{f}_e$}
   \label{fig:eye_blink_ablation}
\end{figure}

\noindent \textbf{Effectiveness of Personalized Attributes.}
Here, we analyze the importance of the disentangled eye blink features and synchronized audio features. As shown in Fig.~\ref{fig:eye_blink_ablation}, the eyes cannot even close without the eye blink features. Table~\ref{tab:audio_ablation} reveals the impact of different audio features. We use LMD and SyncNet confidence scores to measure the accuracy and synchronization of the lip motion. The results show that MFCC and DeepSpeech\cite{deepSpeech} features cannot generate synchronized lip motions accurately. With the help of contrastive learning, our feature can generate much better results.

\noindent \textbf{Impact of VAE Structure and Gaussian Process.} 
We conduct ablation studies on the structure of VAE and the impact of the Gaussian Process. The results are shown in Table~\ref{tab:head_pose_GPVAE_ablation_study}. We use D-Rot/Pos in MakeItTalk\cite{MakeItTalk} to measure head rotation and translation prediction accuracy. It can be observed that both MLP and LSTM cannot achieve satisfactory results, where MLP ignores the temporal relation of the time series and LSTM tends to generate still outputs. Transformer-based architecture improves performance significantly, proving the effectiveness of Transformer on long time series prediction. Additionally, Gaussian Process can capture the temporal dynamics among frames, consequently improving the final performance.
More analysis and experiments can be found in the supplementary material. 

\begin{table}[h]
    \centering
    \begin{tabular}{lcc}
    \toprule
        Mehod & LMD$\,\,\downarrow$ & Sync$_{conf}\,\,\uparrow$  \\
    \midrule
        MFCC & 5.71 & 4.6  \\ 
        Deep Speech~\cite{deepSpeech} & 4.63 & 5.1  \\ 
        Ground Truth & 0.00 & 7.8  \\ 
        Ours & \textbf{3.86} & \textbf{7.3}  \\ 
    \bottomrule
    \end{tabular}
    \caption{Ablation study for the input audio feature in NeRF. We use MFCC features or Deep Speech features as conditions of NeRF. LMD and lip sync scores are adopted to evaluate the lip motion accuracy. Ground Truth refers to the ground truth scores of LMD and Sync$_{conf}$. Our method achieves the best performance.}
    \label{tab:audio_ablation}
\end{table}
\section{Discussion and Conclusion}
\label{sec:conclusion}

We have proposed Disentangled Face Attributes NeRF (DFA-NeRF) for talking head generation. We argue that an excellent talking-head video requires three properties, \ie, maintaining lip-syncing at a semantic level, keeping high visual quality and containing personalized spontaneous motions. DFA-NeRF achieves these properties w.r.t. both personalized attributes and lip motion in talking heads with controllable head pose and eye blink. Moreover, quantitative and qualitative experiments demonstrate that our DFA-NeRF can synthesize high-fidelity talking head videos.

\noindent \textbf{Limitation.} Our method is not applicable for multiple input voices. Studies on speaker diarization splitting up audio into homogeneous segments provide a probable solution. Since we do not focus on fast inference, the rendering process is time-consuming while this can be relieved by acceleration methods.

\noindent \textbf{Broader Impacts.} Our method can generate natural and realistic talking head videos with few requirements. So it may bring potential ethical issues and negative social impacts. Therefore, researchers must treat DFA-NeRF cautiously and rationally. Meanwhile, the community should promote studies on face forge detection to automatically keep these videos away from social media

{\small
\bibliographystyle{ieee_fullname}
\bibliography{egbib}
}

\end{document}